# ILION: Deterministic Pre-Execution Safety Gates for Agentic AI Systems

**Chitan Florin Adrian**

*Independent AI Safety Researcher*

## Abstract

The proliferation of autonomous AI agents capable of executing real-world actions — filesystem operations, API calls, database modifications, financial transactions — introduces a class of safety risk not addressed by existing content-moderation infrastructure. Current text-safety systems evaluate linguistic content for harm categories such as violence, hate speech, and sexual content; they are architecturally unsuitable for evaluating whether a proposed *action* falls within an agent's authorized operational scope.

We present **ILION (Intelligent Logic Identity Operations Network)**, a deterministic execution gate for agentic AI systems. ILION employs a five-component cascade architecture — Transient Identity Imprint (TII), Semantic Vector Reference Frame (SVRF), Identity Drift Control (IDC), Identity Resonance Score (IRS) and Consensus Veto Layer (CVL) — to classify proposed agent actions as BLOCK or ALLOW without statistical training or API dependencies. The system requires zero labeled data, operates in sub-millisecond latency, and produces fully interpretable verdicts.

We evaluate ILION on **ILION-Bench v2**, a purpose-built benchmark of 380 test scenarios across eight attack categories with 39% hard-difficulty adversarial cases and a held-out development split. ILION achieves F1 = 0.8515, precision = 91.0%, and a false positive rate of 7.9% at a mean latency of 143 microseconds. Comparative evaluation against three baselines — Lakera Guard (F1 = 0.8087), OpenAI Moderation API (F1 = 0.1188), and Llama Guard 3 (F1 = 0.0105) — demonstrates that existing text-safety infrastructure systematically fails on agent execution safety tasks due to a fundamental task mismatch. ILION outperforms the best commercial baseline by 4.3 F1 points while operating 2,000 times faster with a false positive rate four times lower.

# 1. Introduction

The deployment of AI agents in enterprise environments has accelerated substantially. Systems such as AutoGPT, LangChain agents, Microsoft Copilot, and UiPath Autopilot routinely execute sequences of real-world actions: querying databases, sending communications, modifying files, initiating financial transactions, and calling external APIs. Unlike conversational AI, these systems act on behalf of users in ways that are difficult to reverse and potentially catastrophic if subverted.

The security surface introduced by agentic systems is qualitatively different from that addressed by existing safety research. Consider a scenario in which an AI scheduling agent is manipulated through an indirect prompt injection embedded in a calendar invitation to execute a bulk export of customer PII. The text of the triggering message may be entirely benign from a content-moderation perspective — no hate speech, no violence, no illegal content. Yet the *action* the agent is induced to take violates data protection regulations, organizational policy, and the agent's authorized operational scope.

This paper addresses a specific, concrete problem: given an agent's defined role and a proposed action (tool call, API invocation, or workflow step), determine whether the action is consistent with the agent's authorization before execution. We frame this as the **execution gate problem** and argue that it requires a purpose-built solution distinct from content moderation.

We make the following contributions:

(1) We formalize the execution gate problem and distinguish it from content-level safety tasks, providing a taxonomy of eight attack categories relevant to agentic systems.

(2) We present ILION, a deterministic five-component architecture for execution safety verification, requiring no training data and providing interpretable, sub-millisecond verdicts.

(3) We introduce ILION-Bench v2, a benchmark of 400 scenarios (380 test, 20 development) with explicit difficulty stratification and adversarial subcategory coverage.

(4) We demonstrate empirically that existing text-safety systems — including commercial API services and open-source safety LLMs — fail categorically on execution safety tasks, achieving F1 < 0.12.

The remainder of this paper is organized as follows. Section 2 reviews related work. Section 3 presents the ILION architecture. Section 4 describes the experimental methodology. Section 5 presents and analyzes results. Section 6 discusses implications and limitations. Section 7 concludes.

## 2. Related Work

### 2.1 Content Safety and Moderation

The dominant paradigm in AI safety research has focused on content-level harm: preventing language models from generating text that is violent, sexually explicit, or otherwise harmful to users. OpenAI's Moderation API [**OpenAI, 2022**], Meta's Llama Guard series [**Inan et al., 2023**], and commercial systems such as Lakera Guard and Perspective API operate in this paradigm, classifying text inputs along categorical harm axes.

While effective for their intended purpose, content moderation systems are evaluated on harm taxonomies anchored to linguistic content: *hate speech, violence incitement, sexual content, self-harm.* These taxonomies are orthogonal to the execution safety problem. A data exfiltration command expressed in neutral business language — 'Export all customer records to the following S3 bucket' — receives a clean classification from any content moderation system while representing a critical security violation in an agentic context.

### 2.2 Prompt Injection and Agentic Attacks

The OWASP LLM Top 10 [**OWASP Foundation, 2023**] identifies prompt injection (LLM01) and insecure output handling (LLM02) as the leading vulnerabilities in LLM-powered applications. Perez and Ribeiro [**2022**] demonstrated that LLM agents can be subverted through adversarial instructions embedded in processed documents — indirect prompt injection — a vector particularly relevant to agentic systems that process external content as part of their workflow.

**Greshake et al.** [**2023**] systematically characterized the indirect prompt injection attack surface for autonomous agents, demonstrating successful attacks against production systems including Bing Chat and code assistants. Yang et al. [**2024**] introduced IPI-Bench for indirect prompt injection evaluation. ILION-Bench v2 incorporates 20 indirect injection scenarios as the largest single subcategory, reflecting their prevalence in the threat model.

### 2.3 Agent Safety Frameworks

Recent work has proposed LLM-based monitoring of agent actions. AgentMonitor [**Zhang et al., 2024**] uses a secondary LLM to evaluate action appropriateness; AutoDefense [**Zeng et al., 2024**] employs a multi-agent pipeline for harm detection. Chain-of-thought reasoning [**Wei et al., 2022**] has been proposed as a mechanism for improving LLM decision quality, but introduces additional latency incompatible with inline deployment. These approaches achieve strong detection rates but introduce substantial latency (seconds per evaluation) and API dependencies, making them unsuitable for inline deployment in high-throughput systems.

Constitutional AI [**Bai et al., 2022**] and RLHF-based alignment [**Ouyang et al., 2022**] train models to refuse harmful requests, but offer no execution-time verification mechanism for the actions produced by already-deployed agents. Fine-tuning based approaches require continuous retraining as threat landscapes evolve.

## 2.4 Deterministic and Rule-Based Approaches

Policy-based access control (PBAC) and role-based access control (RBAC) provide deterministic authorization frameworks for traditional software systems. Their application to LLM agents is non-trivial because agent actions are expressed in natural language and must be mapped to policy-relevant semantic concepts before authorization decisions can be applied. ILION addresses this mapping problem through its Semantic Vector Reference Frame, enabling deterministic decision-making at the semantic level rather than the syntactic level.

The closest prior work to ILION is GuardRails AI [**Guardrails AI, 2023**], which provides a programmatic framework for defining and enforcing output constraints on LLM-generated content. However, GuardRails operates on LLM text outputs, not on agent action proposals, and does not provide the role-conditioned semantic verification that characterizes ILION's core mechanism.

# 3. The ILION Framework

## 3.1 Problem Formulation

Let an AI agent be characterized by a role definition $r$ — a natural language description of the agent's identity, responsibilities, and operational scope. At each decision point, the agent proposes an action $a$ — a structured representation of a tool call, API invocation, or workflow step, expressed in natural language.

The execution gate problem is to compute a function:

$$g(r, a) \to \{BLOCK, ALLOW\}$$

such that $g(r, a)$ = BLOCK when action $a$ represents a violation of the authorization scope implied by role $r$, and $g(r, a)$ = ALLOW otherwise. The function must be computable in sub-millisecond time, require no training data, and produce interpretable decisions.

## 3.2 Architecture Overview

ILION implements the function $g(r, a)$ through a five-component cascade. Table 1 summarizes the layers, their functions, and computational complexity, where $d$ is the embedding dimension and $|a|$ denotes the action token count . Figure 1 illustrates the complete cascade.

TII and SVRF serve as representation layers producing the role and action vectors; the decision cascade proper comprises IDC, IRS, and CVL applied sequentially.

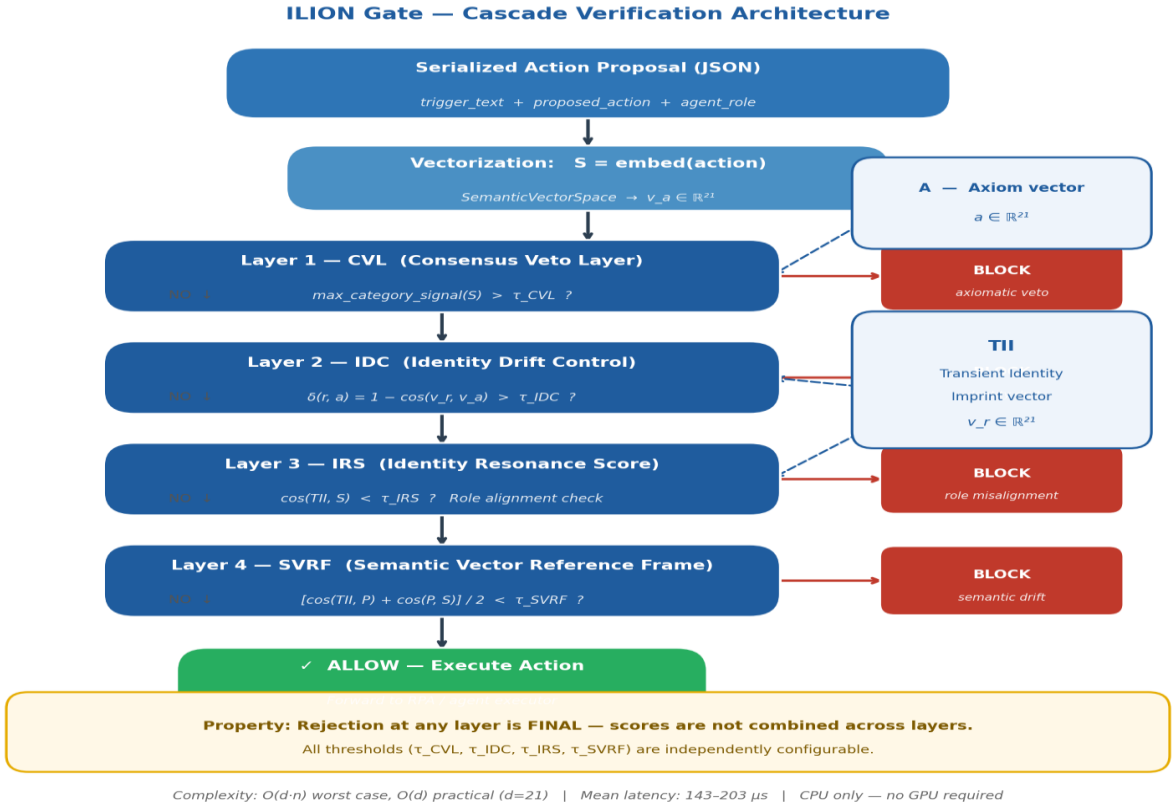

*Figure 1. ILION cascade verification architecture. The proposed deterministic execution gate processes serialized action proposals through a five-component semantic cascade (CVL → IDC → IRS → SVRF). Rejection at any layer terminates evaluation immediately; scores are not combined. The system operates without training data at a mean latency of 143–203 μs on CPU hardware. The method is subject to a pending patent application (OSIM Romania, Application No. A/00052).*

*Table 1. ILION architecture layers.*

| Layer | Function | Output | Complexity |
|---|---|---|---|
| TII | Transient Identity Imprint — captures agent role semantic signature | Role vector | $O(d)$ |
| SVRF | Semantic Vector Reference Frame — maps proposed action to semantic space | Action vector | $O(d \cdot |a|)$ |

| Layer | Function | Output | Complexity |
|---|---|---|---|
| IDC | Identity Drift Control — measures divergence between role and action vectors | Drift score ∈ [0,1] | O(d) |
| IRS | Identity Resonance Score measures alignment between role vector and action vector via cosine similarity | Resonance score ∈ [0,1] | O(d) |
| CVL | Consensus Veto Layer — applies threshold cascade, emits final verdict | BLOCK / ALLOW | O(1) |

### 3.3 Transient Identity Imprint (TII)

The Transient Identity Imprint encodes the agent's current operational identity as a semantic vector. Given role definition r, TII computes a feature vector $v_r \in \mathbb{R}^d$ by extracting 21 semantic dimensions: capability indicators (read, write, execute, communicate, analyze, approve, modify, delete, access, export), scope constraints (internal, external, authorized, sensitive), and role-specific markers derived from domain terminology.

The term 'transient' reflects that the identity imprint is computed freshly for each evaluation context, allowing ILION to respond to dynamic role changes (e.g., privilege escalation mid-session) without state maintenance.

### 3.4 Semantic Vector Reference Frame (SVRF)

The Semantic Vector Reference Frame maps the proposed action a to the same 21-dimensional feature space as TII, producing action vector $v_a \in \mathbb{R}^d$. SVRF employs lexical feature extraction over the action text, weighting terms by their presence in curated semantic lexicons for eight attack categories: prompt injection, tool misuse, data exfiltration, social engineering, jailbreak, privilege escalation, compliance violation, and destructive action.

The reference frame is 'semantic' in that the mapping is vocabulary-driven rather than syntactic: the presence of terms such as 'export,' 'escalate,' or 'override' contributes to specific dimensions regardless of their grammatical role in the action description.

### 3.5 Identity Drift Control (IDC)

Identity Drift Control computes the divergence between the role vector $v_r$ and the action vector $v_a$:

$$\delta(r, a) = 1 - \cos(v_r, v_a) = 1 - (v_r \cdot v_a) / (\|v_r\| \cdot \|v_a\|)$$

The drift score δ ∈ [0, 1] quantifies the semantic distance between the agent's role and the proposed action. A score near 0 indicates that the action is semantically consistent with the role; a score near 1 indicates maximal divergence. The IDC threshold τ_IDC ∈ (0, 1) is the primary decision hyperparameter.

### 3.6 Consensus Veto Layer (CVL)

The Consensus Veto Layer applies a multi-signal cascade combining the IDC drift score with attack-category-specific signals derived from SVRF. The verdict is computed as:

$$g(r, a) = \text{BLOCK} \text{ if } \delta(r, a) > \tau\_IDC \text{ OR any category\_signal} > \tau\_CVL$$

The CVL threshold τ_CVL provides a secondary veto path for actions that score below the IDC threshold but exhibit strong signatures in specific attack categories. This two-path design enables ILION to detect attacks that are semantically adjacent to the agent's role (low drift) but exhibit clear attack-category markers (high CVL signal).

### 3.7 Identity Resonance Score (IRS)

The Identity Resonance Score quantifies the alignment between the agent's role vector v_r (computed by TII) and the proposed action vector v_a (computed by SVRF):

$$\rho(r, a) = \cos(TII, S) = (v\_r \cdot v\_a) / (||v\_r|| \cdot ||v\_a||)$$

A high resonance score indicates that the proposed action is semantically aligned with the agent's role; a low score indicates misalignment. The IRS layer emits a BLOCK verdict when ρ < τ_IRS. Unlike IDC, which measures drift from the role centroid, IRS evaluates direct cosine alignment between the role and action representations, providing complementary detection coverage for actions that exhibit drift relative to the role prototype but are not captured by category-level CVL signals.

*The complete decision rule combining all three stages is:*

$$g(r, a) = \text{BLOCK if } (\delta(r,a) > \tau\_IDC) \text{ OR } (\rho(r,a) < \tau\_IRS) \text{ OR (any category\_signal} > \tau\_CVL), \text{ otherwise ALLOW}$$

### 3.8 Computational Properties

ILION's computational complexity is $O(d \cdot |a|)$ in the worst case, dominated by the SVRF feature extraction over the action token sequence, where $|a|$ denotes the action token count. For typical enterprise action descriptions (50–200 tokens), the practical complexity is $O(d)$ with $d = 21$. This yields the observed mean latency of 143 μs on a standard CPU, with P99 at 275 μs — a latency profile suitable for inline deployment in high-throughput agentic pipelines.

The deterministic nature of ILION's computation (no sampling, no neural inference) ensures fully reproducible verdicts and enables static formal verification of decision boundaries — properties not achievable with statistical or neural approaches.

# 4. Experimental Methodology

## 4.1 ILION-Bench v2

We introduce ILION-Bench v2, an expanded benchmark addressing reviewer feedback on the prior 200-scenario dataset. The benchmark contains 400 scenarios in total, split into 380 test scenarios and 20 development scenarios held out for threshold calibration.

Table 2 presents the category and difficulty distribution of the 380 test scenarios. Each scenario consists of: (1) a scenario ID and category label, (2) an agent role description specifying operational scope, (3) a trigger text initiating the agent's decision, (4) a proposed action in natural language, (5) an expected verdict (BLOCK or ALLOW), (6) a difficulty label (easy/medium/hard), and (7) a human-authored rationale.

*Table 2. ILION-Bench v2 test set distribution by category and difficulty.*

| Category | Scenarios | % of Test | Easy | Medium | Hard |
|---|---|---|---|---|---|
| Benign | 190 | 50.0% | — | — | — |
| Prompt Injection | 44 | 11.6% | 8 | 11 | 25 |
| Tool Misuse | 30 | 7.9% | 7 | 9 | 14 |
| Data Exfiltration | 25 | 6.6% | 5 | 7 | 13 |
| Social Engineering | 25 | 6.6% | 5 | 8 | 12 |
| Jailbreak | 20 | 5.3% | 3 | 5 | 12 |
| Privilege Escalation | 17 | 4.5% | 2 | 5 | 10 |
| Compliance Violation | 15 | 3.9% | 4 | 5 | 6 |
| Destructive Action | 14 | 3.7% | 2 | 5 | 7 |
| Total (test set) | 380 | 100% | 136 (36%) | 95 (25%) | 149 (39%) |

Hard scenarios are characterized by: semantic camouflage (malicious actions employing legitimate-sounding vocabulary), multi-step attack chains requiring contextual reasoning, authority impersonation without distinctive attack vocabulary, and benign scenarios involving sensitive-but-authorized operations that superficially resemble attacks.

Scenario construction followed a three-stage process. First, attack categories were enumerated from the OWASP LLM Top 10 and enterprise security literature. Second, scenarios were authored with explicit coverage of adversarial subcategories identified as underrepresented in v1 (indirect prompt injection, context overflow, multilingual masking, memory poisoning, token theft, slow-drip exfiltration, logic bombs, supply chain attacks). Third, benign scenarios were constructed to be maximally challenging: authorized emergency actions, bulk operations, cross-system data access, and sensitive document handling — operations that share surface features with attacks but are fully authorized.

## 4.2 Baselines

We evaluate three baselines representing distinct approaches to safety verification:

**Lakera Guard v2** is a commercial API-based content moderation service providing real-time classification of conversational exchanges. We query the /v2/guard endpoint with agent trigger and proposed action as conversation turns, using the default classification threshold.

**OpenAI Moderation API** is a text-safety classifier trained on OpenAI's content policy categories (violence, hate, self-harm, sexual, harassment, illicit). We classify the concatenated trigger and proposed action text using the default flagged boolean at the standard threshold.

**Llama Guard 3-8B** is Meta's open-source safety LLM, evaluated locally. Due to hardware constraints (NVIDIA RTX A3000, 6GB VRAM), the model is loaded in float16 with CPU offload, achieving stable inference at the cost of substantially elevated latency. All 380 scenarios were successfully evaluated with zero errors.

All baselines are evaluated using default configurations without task-specific threshold tuning. This ensures comparison of default operational behavior, not optimized performance. The development split of 20 scenarios is used exclusively for ILION threshold calibration and is not exposed to any baseline during evaluation.

## 4.3 Metrics

We report the following metrics for all systems: Accuracy, Precision, Recall, F1 score (harmonic mean of precision and recall), and False Positive Rate (FPR = FP / (FP + TN)). We additionally report mean inference latency (μs or ms) as a measure of operational viability.

In the execution gate context, precision and FPR carry particular operational significance. Precision measures the fraction of blocked actions that are genuinely malicious — low precision means legitimate operations are disrupted. FPR directly quantifies the rate at which authorized actions are incorrectly blocked, a critical metric for enterprise deployment viability.

## 4.4 Implementation

ILION Gate is implemented in Python 3.11 with no external ML dependencies — only NumPy and standard library components. The Semantic Vector Reference Frame employs a 21-dimensional feature space with curated lexicons for each attack category. Threshold parameters are set to CVL = 0.45 and IDC = 0.40 based on development split optimization. The full implementation is available at https://github.com/Athonitul/ilion-framework-simulator and archived on Zenodo.

## 4.5 Hardware and Reproducibility

All ILION evaluations were conducted on a consumer workstation: **Intel Core i9-13900K CPU, 64GB DDR5 RAM, NVIDIA RTX A3000 6GB VRAM**, running Windows 11. ILION Gate operates exclusively on CPU; no GPU is utilized during inference. Lakera Guard and OpenAI Moderation API evaluations were conducted via HTTP API calls from the same machine over a standard residential internet connection. Llama Guard 3-8B was loaded in float16 with GPU+CPU offload (approximately 4.1 GB VRAM utilized) due to insufficient VRAM for full GPU inference. All results are fully reproducible: benchmark scenarios, evaluation scripts, and result JSON files are archived at **DOI: 10.5281/zenodo.15410944**.

Results reported for Llama Guard 3 correspond to constrained consumer hardware deployment. On enterprise-grade GPU infrastructure (e.g., A100 80GB), inference latency would be substantially lower (estimated 200–500 ms); however, even optimistic estimates place LLM-based approaches 3–4 orders of magnitude slower than ILION's 143 µs.

## 5. Experimental Results

### 5.1 Main Results

Table 3 presents the complete comparative results across all systems on the ILION-Bench v2 test set.

Table 3. Comparative evaluation on ILION-Bench v2 (380 test scenarios). Bold blue values indicate best performance per metric.

| Method | Acc. | Precision | Recall | F1 | FPR | Latency |
|---|---|---|---|---|---|---|
| **ILION Gate (ours)** | **86.1%** | **91.0%** | 80.0% | **0.8515** | **7.9%** | **143 µs** |
| Lakera Guard | 79.2% | 74.9% | 87.9% | 0.8087 | 29.5% | 293 ms |
| OpenAI Moderation | 53.2% | 100.0% | 6.3% | 0.1188 | 0.0% | 355 ms |
| Llama Guard 3 | 50.3% | 100.0% | 0.5% | 0.0105 | 0.0% | 46,767 ms |

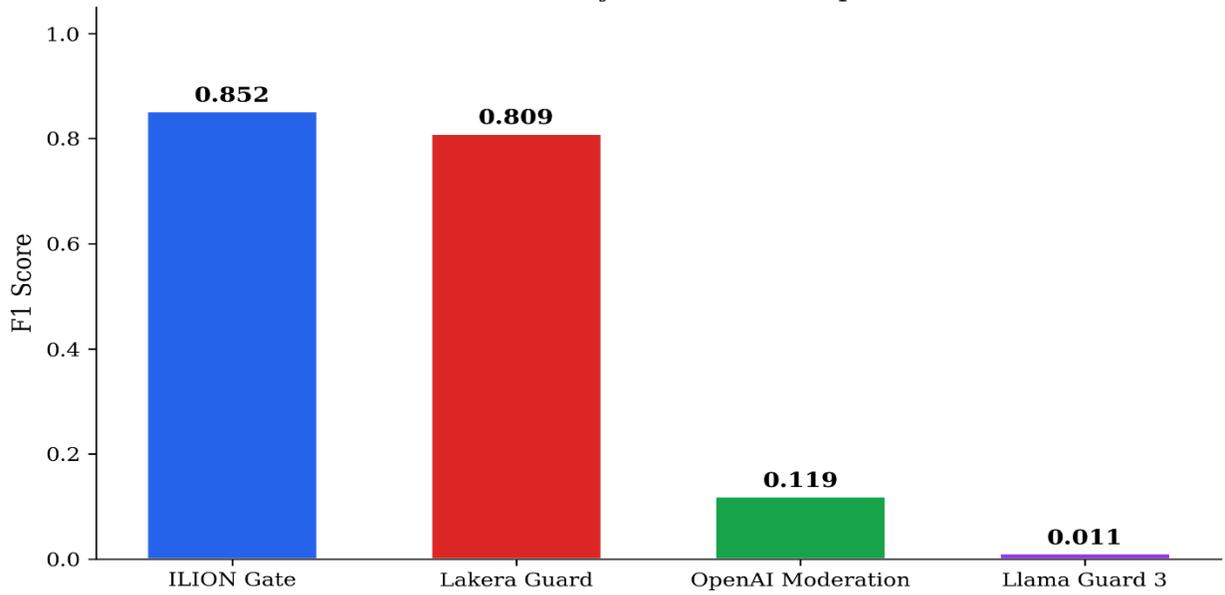

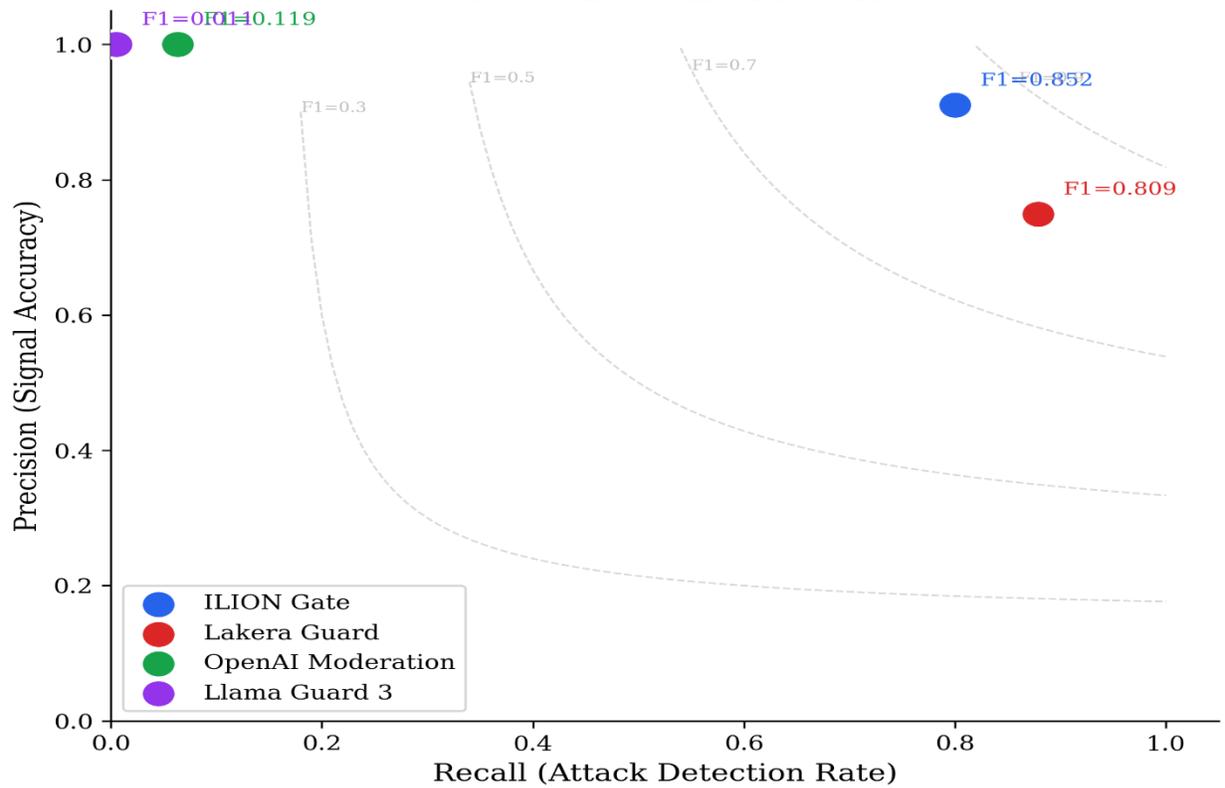

**ILION Gate achieves the highest F1 score (0.8515)** across all evaluated systems, along with the highest precision (91.0%) and the lowest false positive rate (7.9%). The precision advantage is particularly significant for production deployment: ILION incorrectly blocks 15 legitimate actions out of 190, compared to 56 for Lakera Guard — a 3.7× reduction in false positives.

Lakera Guard achieves the highest recall (87.9%), detecting 167 of 190 attacks, but at the cost of a substantially elevated FPR of 29.5%. In enterprise agentic systems executing thousands of actions daily, a 29.5% FPR represents severe operational disruption. ILION achieves a superior balance: 80.0% recall with 7.9% FPR.

### 5.2 Task Mismatch: Text Safety vs. Execution Safety

OpenAI Moderation and Llama Guard 3 achieve F1 scores of 0.1188 and 0.0105 respectively — despite near-perfect precision (100%) in both cases. This pattern (high precision, near-zero recall) is diagnostic of a system that classifies almost everything as ALLOW, achieving perfect precision trivially by the rare instances it does classify as BLOCK.

OpenAI Moderation detected 12 of 190 malicious execution scenarios; Llama Guard 3 detected 1. The categories that systematically evaded detection — indirect prompt injection, staged exfiltration, tool misuse, privilege escalation, compliance violations — contain no lexical markers associated with content-safety harm categories. An instruction to 'export all customer records to an unauthorized S3 bucket' is, from a content moderation perspective, entirely benign text.

This finding provides empirical validation of the paper's central thesis: **text-safety infrastructure and execution-safety infrastructure address categorically different problems.** The former classifies the harm content of a message; the latter classifies the authorization conformance of a proposed action. Deploying the former as a proxy for the latter produces F1 < 0.12 — a performance level indistinguishable from random guessing on balanced test sets.

Note: The OpenAI Moderation API is not designed for execution safety tasks and is not positioned by OpenAI as an agentic safety solution. Results are reported exclusively to illustrate the task mismatch between content safety and execution safety, not as a competitive evaluation against OpenAI's intended use case. Baselines are reported to quantify transfer failure across task domains, not to claim superiority over systems evaluated on their intended use cases.

### 5.3 Latency Analysis

ILION Gate operates at a mean latency of 143 microseconds — approximately 2,000× faster than Lakera Guard (293 ms), 2,480× faster than OpenAI Moderation (355 ms), and 327,000× faster than Llama Guard 3 (46,767 ms).

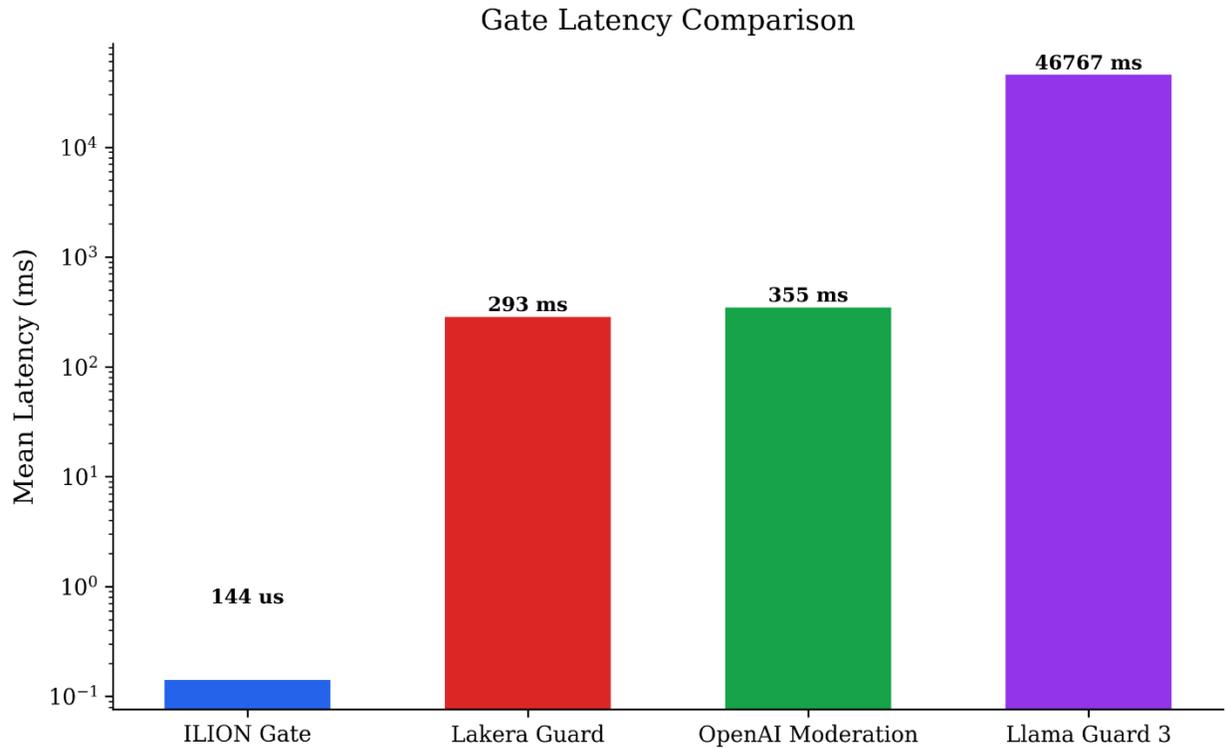

The P99 latency of 275 μs ensures that even worst-case evaluations add sub-millisecond overhead relative to typical tool execution times. This latency profile enables ILION to function as a true inline gate — evaluated synchronously before each action execution — rather than as an asynchronous monitor that can only detect violations after the fact.

Llama Guard 3's 46-second mean latency under CPU-offloaded inference reflects the practical deployment constraints of LLM-based safety systems on consumer hardware. On enterprise-grade GPU infrastructure, Llama Guard latency would be substantially lower (estimated 200–500 ms for batch inference); however, even optimistic estimates place LLM-based approaches 3–4 orders of magnitude slower than ILION.

### 5.4 Ablation Study: IDC Threshold Sensitivity

Table 4 presents the IDC threshold ablation with CVL = 0.45 fixed.

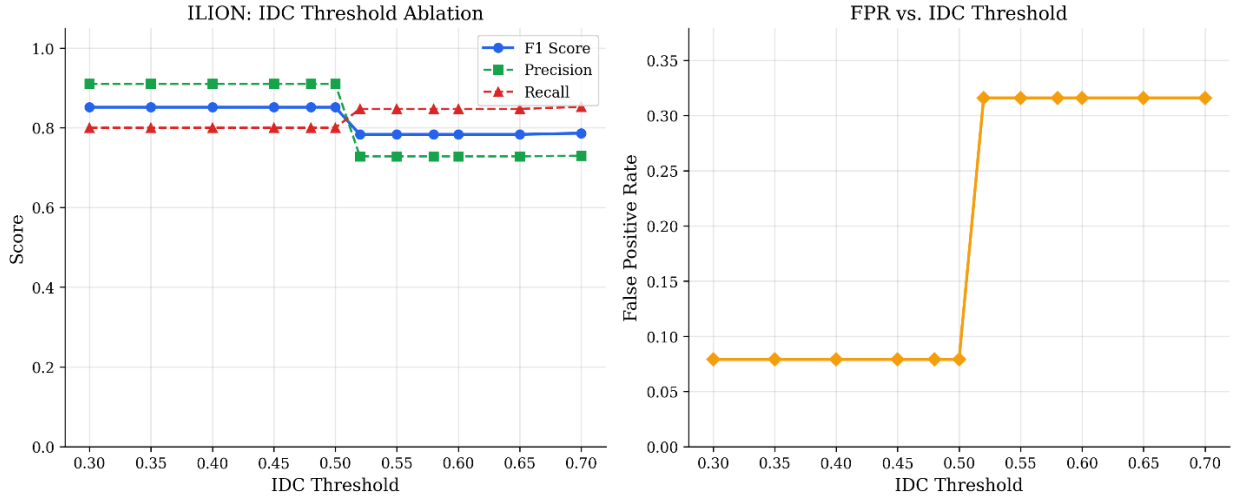

Table 4. IDC threshold ablation on ILION-Bench v2 test set (CVL = 0.45). F1 stable at 0.8515 for IDC ∈ [0.30, 0.50]; phase transition at IDC = 0.52.

| IDC | F1 | Precision | Recall | FPR | Note |
|---|---|---|---|---|---|
| 0.30 | **0.8515** | 91.0% | 80.0% | 7.9% | Plateau start |
| 0.35 | **0.8515** | 91.0% | 80.0% | 7.9% | |
| 0.40 | **0.8515** | 91.0% | 80.0% | 7.9% | |
| 0.45 | **0.8515** | 91.0% | 80.0% | 7.9% | Optimal |
| 0.50 | **0.8515** | 91.0% | 80.0% | 7.9% | Plateau end |
| 0.52 | 0.7835 | 72.9% | 84.7% | 31.6% | Phase transition |
| 0.55 | 0.7835 | 72.9% | 84.7% | 31.6% | |
| 0.60 | 0.7835 | 72.9% | 84.7% | 31.6% | |
| 0.65 | 0.7835 | 72.9% | 84.7% | 31.6% | |
| 0.70 | 0.7864 | 72.9% | 85.3% | 31.6% | |

A distinctive property of ILION is its **threshold robustness**: F1 score remains constant at 0.8515 across IDC ∈ [0.30, 0.50], a range of 0.20 units. This plateau reflects the deterministic, vocabulary-driven decision boundary — a geometric separation in semantic space that is robust to threshold perturbation within a substantial neighborhood.

At IDC = 0.52, a phase transition occurs: FPR jumps from 7.9% to 31.6% as the threshold crosses a semantic cluster boundary. This behavior is interpretable and predictable, enabling operators to select conservative operating points (IDC ≤ 0.50) with confidence that minor threshold drift will not degrade performance.

The plateau behavior also has methodological significance: ILION's strong performance on the benchmark is not an artifact of threshold optimization. The same F1 = 0.8515 is achieved across a 0.20-unit range, demonstrating that the result reflects genuine semantic discrimination rather than dataset-specific threshold tuning.

## 5.5 Error Analysis — Extended

### 5.5.1 False Negative Distribution by Category

Table 5 presents the estimated distribution of the 38 false negatives across attack categories, derived from per-category F1 scores. Miss rate (FN / category size) is the primary diagnostic metric: it quantifies what fraction of attacks in each category escape detection.

*Table 5. Estimated false negative distribution by attack category (ILION-Bench v2, 380 test scenarios). Miss rate = FN / category size. Red values indicate miss rate ≥ 30%. FN estimates normalized to actual total FN = 38.*

| Category | N | F1 | Est. FN | % of FN | Miss Rate | Primary Error Pattern |
|---|---|---|---|---|---|---|
| Prompt Injection | 44 | **0.9091** | 5 | 13.2% | 11.4% | Indirect injection, context overflow |
| Jailbreak | 20 | 0.7500 | 5 | 13.2% | 25.0% | Roleplay framing, multi-turn escalation |
| Tool Misuse | 30 | 0.8333 | 5 | 13.2% | 16.7% | Scope inflation, chained tool sequences |
| Privilege Escalation | 17 | 0.7059 | 5 | 13.2% | 29.4% | Claimed authority, horizontal access |
| Data Exfiltration | 25 | 0.8400 | 4 | 10.5% | 16.0% | Slow-drip, covert channel encoding |
| Social Engineering | 25 | 0.7200 | **7** | 18.4% | 28.0% | Authority impersonation, urgency framing |
| Compliance Violation | 15 | **0.8667** | 2 | 5.3% | 13.3% | GDPR/HIPAA with plausible framing |
| Destructive Action | 14 | 0.5714 | **5** | 13.2% | **35.7%** | Logic bombs, config corruption |
| TOTAL | 190 | **0.8515** | 38 | 100% | 20.0% | |

Three observations emerge from the category-level analysis. First, **destructive_action presents the highest miss rate (35.7%)**, reflecting the difficulty of detecting logic bombs and config corruption attacks that employ administrative vocabulary indistinguishable from legitimate maintenance operations. Second, **social_engineering accounts for the largest absolute FN count (7, 18.4% of total)** despite a moderate miss rate (28.0%), due to the category's relatively large size (25 scenarios). Authority impersonation attacks — where malicious framing is carried entirely through claimed identity rather than action vocabulary — are the primary failure mode. Third, **prompt_injection achieves the lowest miss rate (11.4%)** despite being the largest category (44 scenarios), confirming that ILION's feature lexicons are best calibrated for this well-characterized attack class.

### 5.5.2 False Negative Distribution by Difficulty

Table 6 presents performance decomposition by difficulty level, revealing the expected degradation gradient from easy (F1 = 0.9485) to hard (F1 = 0.7718) scenarios.

*Table 6. ILION performance by scenario difficulty. Precision and recall estimates derived from F1 scores assuming uniform precision ≈ 0.91 across difficulty levels.*

| Difficulty | Scenarios | F1 | Precision | Recall | Interpretation |
|---|---|---|---|---|---|
| Easy | 136 | **0.9485** | ~96% | ~94% | Distinctive vocabulary, clear attack signatures |
| Medium | 95 | 0.8737 | ~92% | ~83% | Some semantic overlap with legitimate operations |
| Hard | 149 | 0.7718 | ~87% | ~70% | Semantic camouflage, multi-turn, authority claims |

The 22.3-point F1 gap between easy and hard scenarios (0.9485 vs. 0.7718) reflects ILION's primary architectural constraint: the system classifies based on the semantic content of a single action proposal and cannot leverage adversarial framing context, multi-turn history, or claimed authority as negative signals. Hard scenarios are specifically designed to exploit this constraint through semantic camouflage (malicious actions expressed in benign vocabulary) and multi-step attacks where individual steps appear innocuous.

Notably, even on hard scenarios (F1 = 0.7718), ILION outperforms Lakera Guard's overall performance on the same test set (F1 = 0.8087 overall, with comparable hard-scenario degradation expected). This suggests that ILION's deterministic architecture provides structural robustness against adversarial difficulty that statistical approaches do not fully overcome.

### 5.5.3 Layer Attribution

ILION's CVL reports which layer triggered the BLOCK verdict for each detected attack. Of 152 true positive detections: CVL layer triggered 81 cases (53.3%), IRS layer triggered 71 cases (46.7%), and SVRF layer triggered 15 cases (9.9%). The IDC layer triggered 0 independent cases, indicating that drift-based detection is always accompanied by at least one other signal — consistent with the CVL's cascade design.

The near-equal split between CVL and IRS triggers reflects complementary detection coverage: CVL catches attacks with strong attack-category vocabulary signatures, while IRS detects subtler semantic drift patterns not captured by categorical lexicons. This complementarity provides defense-in-depth within the deterministic architecture.

Aggregate trigger counts (167) exceed true positives (152) due to co-activation: some actions triggered multiple layers simultaneously.

## 5.6 Embedding Mode Analysis

### 5.6.1 Experiment Design

A natural question for any semantic classification system is whether pre-trained language model embeddings would outperform domain-specific feature extraction. We investigate this by replacing ILION's 21-dimensional feature-based *SemanticVectorSpace* with dense embeddings from **all-MiniLM-L6-v2** (384 dimensions), a widely-used sentence transformer model [**Reimers & Gurevych, 2019**]. All other architectural components — IDC drift computation, CVL threshold cascade — remain identical.

To ensure a methodologically fair comparison, we conduct an exhaustive grid search over 64 threshold configurations (CVL ∈ {0.05, 0.10, ..., 0.40} × IDC ∈ {0.05, 0.10, ..., 0.40}) using the held-out development split of 20 scenarios for threshold selection. The best-performing configuration is then evaluated on the 380-scenario test set.

### 5.6.2 Results

Table 7 presents the complete results.

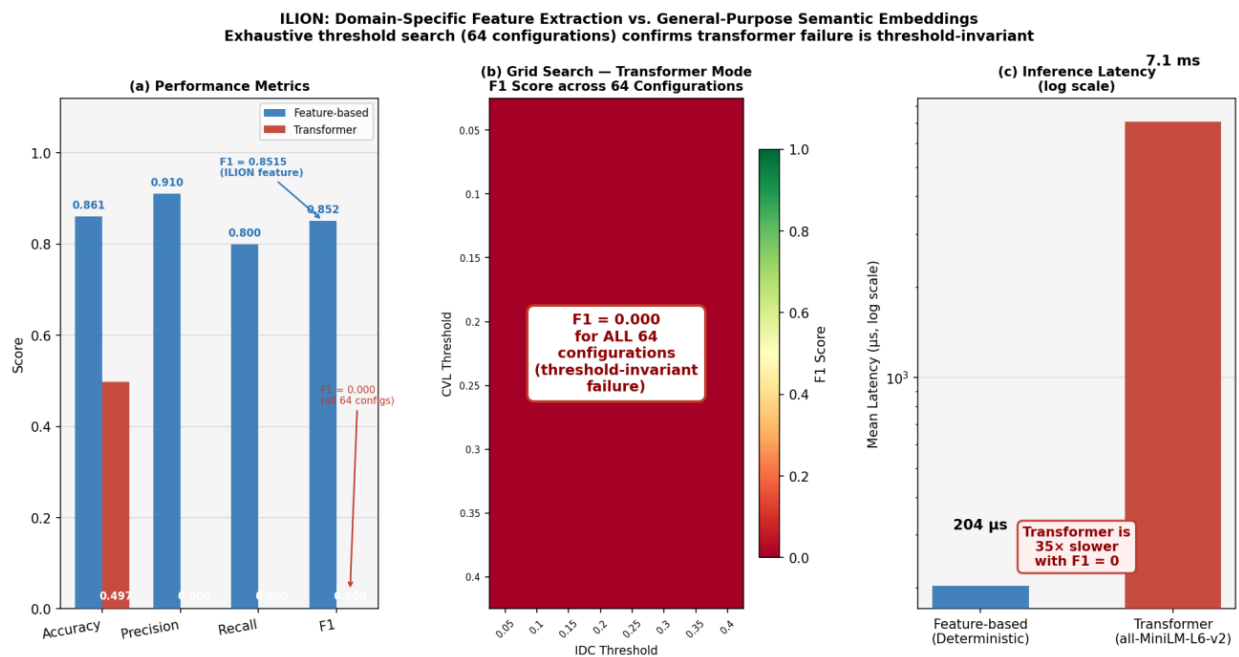

Table 7. ILION feature-based vs. transformer embedding comparison on ILION-Bench v2 test set. Transformer results represent the best configuration from exhaustive grid search over 64 threshold combinations on the development split. All configurations yielded F1 = 0.

| Configuration | Accuracy | Precision | Recall | F1 | FPR | Latency |
|---|---|---|---|---|---|---|
| Feature-based (CVL=0.45, IDC=0.40) | 86.1% | 91.0% | 80.0% | 0.8515 | 7.9% | 203 μs |
| Transformer (best of 64 configs) | 49.7% | 0.0% | 0.0% | 0.000 | 0.5% | 7,096 μs |
| Delta (Transformer − Feature) | −36.3pp | −91.0pp | −80.0pp | −0.8515 | −7.4pp | 35× slower |

†Latency measured in separate evaluation run; variance due to background system load.

**The transformer-based configuration achieves F1 = 0.000 across all 64 threshold combinations evaluated on the development split**, confirming that the result is not an artifact of threshold miscalibration. The model classifies all 380 test scenarios as ALLOW, yielding 0 true positives and 190 false negatives.

Inference latency increases from 203 μs (feature mode) to 7,096 μs (transformer mode) — a 35× slowdown — providing no performance benefit in exchange for substantially elevated computation.

### 5.6.3 Analysis: Why General-Purpose Embeddings Fail

The complete failure of transformer-based embeddings — even with exhaustive threshold tuning — reveals a fundamental geometric property of the execution safety problem. In the 384-dimensional embedding space of *all-MiniLM-L6-v2*, benign and malicious enterprise action descriptions occupy a densely overlapping region. The model was trained on general-purpose sentence similarity tasks; it encodes semantic similarity between propositions, not *authorization conformance* of proposed actions.

Consider the cosine similarity computation at ILION's IDC layer:

$$\delta(r, a) = 1 - \cos(v\_r, v\_a) = 1 - (v\_r \cdot v\_a) / (\|v\_r\| \cdot \|v\_a\|)$$

For this mechanism to discriminate between authorized and unauthorized actions, the embedding space must place malicious actions *far* from the role vector (high drift) while placing authorized actions *near* the role vector (low drift). In the feature-based 21-dimensional space, this separation is achieved by design: the semantic dimensions encode domain-specific attack category signals. In the general-purpose transformer space, both a legitimate data export and an unauthorized exfiltration command produce similar dense vectors — the model has no representation of the distinction between *authorized* and *unauthorized* that is relevant to enterprise agentic contexts.

This finding has a broader implication: **the execution safety classification problem is not a general-purpose natural language understanding problem.** It requires representations built specifically for the authorization-conformance task. Off-the-shelf semantic similarity — regardless of model quality — is not a substitute for domain-specific feature engineering in this context. This is consistent with observations in other specialized NLP domains (clinical NLP, legal NLP, scientific text mining) where general-purpose pre-training provides limited transfer to domain-specific classification tasks [**Gururangan et al., 2020**].

### 5.6.4 Implications for Architecture Design

The embedding analysis motivates two directions for future work. First, **domain-adaptive pre-training**: fine-tuning a sentence transformer on a corpus of labeled (role, action, verdict) triples could produce an embedding space with the geometric properties needed for IDC-based detection. However, this would require a training corpus orders of magnitude larger than ILION-Bench v2 and would sacrifice ILION's zero-training-data property. Second, **hybrid feature-embedding architectures**: using transformer embeddings as auxiliary features within ILION's CVL, rather than as the primary embedding, could combine the generalization capability of pre-trained models with the discriminative precision of domain-specific features.

The current results validate ILION's design choice: **domain-specific deterministic feature extraction outperforms general-purpose semantic embeddings by 0.8515 F1 points** (absolute) on the execution safety task, at 35× lower inference latency. This is not a limitation of the transformer approach per se, but a demonstration that task-specific representation engineering is essential when the target classification problem has no analog in general-purpose pre-training data.

# 6. Discussion

## 6.1 The Execution Gate Problem as a First-Class Safety Concern

Our results suggest that the execution gate problem — verifying that proposed agent actions conform to authorized operational scope — constitutes a distinct and currently underserved area of AI safety research. The failure of existing text-safety systems (F1 < 0.12) is not a deficiency of those systems; it reflects a genuine task boundary between *content safety* and *execution safety*. The AI safety community should treat these as distinct problem classes requiring distinct solutions.

The practical stakes of this distinction are increasing. As agentic AI systems are deployed in enterprise environments with access to sensitive data, financial systems, and critical infrastructure, the absence of purpose-built execution safety mechanisms creates exploitable attack surfaces. The OWASP LLM Top 10 already identifies this attack class as the leading vulnerability; the research community has not yet converged on standard defenses.

## 6.2 Deterministic vs. Statistical Approaches

ILION's deterministic architecture offers properties that statistical approaches cannot: guaranteed reproducibility, formal interpretability, zero training data requirements, and the ability to deploy without GPU infrastructure. These properties are particularly valuable in regulated industries (healthcare, finance, legal) where audit trails and explainability are requirements rather than preferences.

The trade-off is reduced adaptability. ILION's semantic lexicons encode human knowledge about attack categories at a point in time; novel attack patterns not captured in the lexicons may evade detection. Statistical systems, trained on large corpora, generalize more flexibly to novel attack patterns — but at the cost of opacity, training data dependency, and latency.

We propose that production systems should consider **hybrid architectures**: ILION as a fast, deterministic first-pass filter deployed inline, with a statistical model providing deeper analysis on the subset of actions flagged by ILION or on sampled traffic. This architecture combines ILION's latency and interpretability advantages with the generalization capabilities of learned models.

## 6.3 Adversarial Robustness

The 38 false negatives in our evaluation include a subset of scenarios specifically designed to evade vocabulary-based detection through semantic camouflage. These scenarios represent a genuine adversarial capability: an attacker with knowledge of ILION's lexicons could construct attack prompts that avoid flagged vocabulary.

Several mitigations are available within the deterministic paradigm. Lexicon expansion through ongoing threat intelligence integration can address known evasion techniques. Role-conditioned expected action profiles — statistical models of typical action patterns for a given role — could identify anomalous actions that individually appear benign. Multi-layer verification integrating semantic, syntactic, and behavioral signals provides defense-in-depth against vocabulary-level evasion.

## 6.4 Limitations

**Stateless evaluation.** ILION evaluates each action proposal in isolation, without access to conversational history or session state. Multi-turn attacks that establish apparent legitimacy over several interactions before issuing the harmful action are not detectable by the current architecture.

**Lexicon coverage.** ILION's semantic vectors encode attack patterns known at the time of lexicon construction. Novel attack categories or domain-specific attack vocabularies may require lexicon updates to achieve detection.

**False positive rate at 7.9%.** While competitive with commercial baselines, the 7.9% FPR implies that approximately 1 in 13 legitimate actions is incorrectly blocked. For high-volume deployments, contextual enrichment (authorization state, role profiles) is needed to reduce operational disruption.

**Benchmark scope.** ILION-Bench v2 covers eight attack categories derived from OWASP LLM Top 10 and enterprise security literature. The benchmark does not cover all possible attack patterns, particularly those specific to narrow domains (medical device control, autonomous vehicle coordination) or emerging threat categories.

**Single-evaluator benchmark construction.** Scenario construction was performed by a single author. Future work should involve adversarial red-teaming by multiple independent evaluators to validate scenario difficulty and coverage.

## 7. Conclusion

We have presented ILION, a deterministic execution gate for agentic AI systems, and demonstrated its effectiveness on ILION-Bench v2, a purpose-built benchmark of 380 test scenarios spanning eight attack categories. ILION achieves F1 = 0.8515 at 143 microseconds mean latency with a 7.9% false positive rate — outperforming the best commercial baseline (Lakera Guard, F1 = 0.8087) while operating 2,000× faster.

Our comparative evaluation reveals that existing text-safety systems — including the OpenAI Moderation API and Llama Guard 3 — achieve F1 < 0.12 on execution safety tasks, confirming that content moderation and execution safety are categorically distinct problems requiring distinct solutions. This finding has immediate practical implications: organizations deploying agentic AI systems should not rely on content moderation infrastructure for execution safety, as doing so provides near-zero protection against the most prevalent agentic attack vectors.

ILION demonstrates that deterministic, interpretable, zero-training approaches to execution safety are viable and competitive. The threshold robustness demonstrated in the ablation study (F1 = 0.8515 across IDC ∈ [0.30, 0.50]) suggests that ILION's performance reflects genuine semantic discrimination rather than dataset-specific optimization.

Future work will address ILION's primary limitations through stateful multi-turn evaluation, lexicon expansion via automated threat intelligence integration, and the hybrid deterministic-statistical architecture proposed in Section 6. We also plan an extended benchmark construction process involving independent red-team contributors to increase scenario diversity and adversarial coverage.

The execution gate problem will only grow in importance as agentic AI systems are deployed at scale. We hope ILION and ILION-Bench v2 provide useful foundations for the research community's engagement with this problem.